\documentclass[letterpaper, 10 pt, conference]{ieeeconf}  

\IEEEoverridecommandlockouts                              

\usepackage{amssymb}
\usepackage{amsmath}
\usepackage{acro}
\usepackage{pdfpages}
\usepackage{subcaption}
\usepackage{tabularx}
\usepackage{float}
\usepackage{cite}
\usepackage{gensymb}
\usepackage{algorithm, algorithmic}
\usepackage{booktabs}
\usepackage{multirow}
\usepackage{graphbox}

\usepackage{hhline}
\usepackage{todonotes}
\usepackage{siunitx}
\usepackage{balance}
\usepackage{cite}

\DeclareMathOperator*{\argmax}{argmax} 
\DeclareAcronym{CPP}{
short = CPP ,
long = coverage path planning
}
\DeclareAcronym{DH}{
short = DH ,
long = data harvesting
}
\DeclareAcronym{UAV}{
short = UAV ,
long = unmanned aerial vehicle
}
\DeclareAcronym{RL}{
short = RL ,
long = reinforcement learning
}
\DeclareAcronym{DQN}{
short = DQN ,
long = deep Q-network
}
\DeclareAcronym{DDQN}{
short = DDQN ,
long = double deep Q-network
}
\DeclareAcronym{NN}{
short = NN ,
short-plural = NNs ,
long = neural network ,
long-plural = neural networks
}
\DeclareAcronym{FoV}{
short = FoV ,
long = field of view
}
\DeclareAcronym{TD}{
short = TD ,
long = temporal difference
}

\title{\LARGE \bf
 UAV Path Planning using Global and Local Map Information with Deep Reinforcement Learning
}

\author{Mirco Theile$^{1}$, Harald Bayerlein$^{2}$, Richard Nai$^{1}$, David Gesbert$^{2}$, and  Marco Caccamo$^{1}$%
\thanks{$^{1}$Mirco Theile, Richard Nai, and Marco Caccamo  are  with the TUM School of Engineering and Design, Technical University of Munich, Germany
        {\tt\small \{mirco.theile, richard.nai, mcaccamo\}@tum.de}}%
\thanks{$^{2}$Harald Bayerlein and David Gesbert are with the Communication Systems Department, \mbox{EURECOM}, Sophia Antipolis, France
        {\tt\small \{harald.bayerlein, david.gesbert\}@eurecom.fr}}%
}

\begin{document}

\maketitle
\thispagestyle{empty}
\pagestyle{empty}

\begin{abstract}

Path planning methods for autonomous unmanned aerial vehicles (UAVs) are typically designed for one specific type of mission. This work presents a method for autonomous UAV path planning based on deep reinforcement learning (DRL) that can be applied to a wide range of mission scenarios. Specifically, we compare coverage path planning (CPP), where the UAV's goal is to survey an area of interest to data harvesting (DH), where the UAV collects data from distributed Internet of Things (IoT) sensor devices. By exploiting structured map information of the environment, we train double deep Q-networks (DDQNs) with identical architectures on both distinctly different mission scenarios to make movement decisions that balance the respective mission goal with navigation constraints. By introducing a novel approach exploiting a compressed global map of the environment combined with a cropped but uncompressed local map showing the vicinity of the UAV agent, we demonstrate that the proposed method can efficiently scale to large environments. We also extend previous results for generalizing control policies that require no retraining when scenario parameters change and offer a detailed analysis of crucial map processing parameters' effects on path planning performance.

\end{abstract}
\section{Introduction}

Autonomous unmanned aerial vehicles (UAVs) are envisioned for a multitude of applications that all require efficient and safe path planning methods, which necessitate the combination of a mission goal with navigation constraints, e.g., flying time and obstacle avoidance. Examples for these applications are area coverage path planning (CPP) \cite{Theile2020}, and data harvesting (DH) from Internet of Things (IoT) sensor nodes \cite{Bayerlein2020}. As its name suggests, covering all points inside an area of interest with CPP is related to conventional path planning, where the goal is to find a path between start and goal positions. In general, CPP aims to cover as much of the target area as possible within given energy or path-length constraints while avoiding obstacles or no-fly zones. 

In the DH scenario, the UAV's goal is to collect data from IoT devices distributed in an urban environment, which implies challenging radio channel conditions through alternating line-of-sight (LoS) and non-line-of-sight (NLoS) links between UAV and IoT devices through building obstruction. DH and CPP are very similar when described as an RL problem since the path planning problem's constraints are mostly identical, and only the goal function changes. In previous work, we have looked at CPP \cite{Theile2020} and DH \cite{Bayerlein2020} separately. We show that both problems can be solved using the same deep reinforcement learning (DRL) approach based on feeding spatial map information directly to the DRL agent via convolutional network layers. This work's focus lies in proposing improvements to existing DRL approaches to generalized, large-scale UAV path planning problems with CPP and DH as examples.

Using maps as a direct input becomes problematic for large map sizes, as the network's size, trainable parameters, and training time increase equivalently. We introduce a global-local map scheme to address the scalability problems of the standard map-based input. In path planning, the intuition is that distant features lead to general direction decisions, while close features lead to immediate actions such as collision avoidance. Thus, the level of detail passed to the agent for distant objects can be less than for close objects. With the global map, a compressed version of the complete environment map centered on the agent's position, general information of all objects on the map is provided to the agent. In contrast, the local map, uncompressed but cropped to show only the UAV agent's immediate surroundings, provides detailed local information.

While numerous path planning algorithms for both problems exist, DRL offers the possibility to solve both distinctly different problems with the same approach. For each problem, DRL agents can learn control policies that generalize over a large scenario parameter space requiring no expensive retraining or recomputation when the scenario changes. However, previous work usually only focuses on finding optimal paths for one single scenario at a time. The DRL paradigm is popular in this context because of its flexibility regarding prior knowledge and assumptions about the environment, the computational efficiency of DRL inference, as well as the complexity of autonomous UAV control tasks, which are usually non-convex optimization problems and proven to be NP-hard in many instances \cite{Zeng2019, Shakeri2019}. A general summary of issues in using UAVs as part of communication networks, including IoT data harvesting, can be found in \cite{Zeng2019}. A survey of various applications for UAV systems from a cyber-physical perspective is offered in \cite{Shakeri2019}. Cabreira \textit{et al.} \cite{Cabreira2019} provide a survey of UAV coverage path planning.

Previous works in UAV path planning have already made use of convolutional map processing for DRL agents. In the drone patrolling problem presented in \cite{Piciarelli2019}, a local relevance map of the patrolling area showing the agent's vicinity cropped to a fixed size is fed into a DDQN agent. No information of the physical environment is included, and there is no consideration for navigation constraints like obstacle avoidance or flying time. In \cite{Julian2019}, fixed-wing UAVs are tasked with monitoring a wildfire propagating stochastically over time. Control decisions are based on either direct observations or belief maps fed into the DRL agents. The focus here is the inherent uncertainty of the problem, not balancing a mission goal and navigation constraints in large complex environments. Wildfire surveillance is also the mission of the quadcopter UAVs in \cite{Seraj2020}, which is set in a similar scenario without navigation constraints and makes use of uncertainty maps to guide path planning. Their approach is based on an extended Kalman filter and not the reinforcement learning (RL) paradigm. To monitor another natural disaster situation, Baldazo \textit{et al.} \cite{Baldazo2019} present a multi-agent DRL method for flood surveillance using the UAVs' local observations of the inundation map to make control decisions. All mentioned approaches focus on solving a single class of UAV missions in simple physical environments and do not consider combining local and global map information.

Missions, where UAVs provide communication services to ground users or devices, include the work in \cite{Esrafilian2018} set in a complex urban environment where the UAV path planning is based on exploiting map information with a method combining dynamic and sequential convex programming. In \cite{Liu2021}, data is collected simultaneously with ground and aerial vehicles on a small map with obstacles. Due to the small map size, the full global map information can be fed into the DRL agents. Another scenario is investigated in \cite{Zhang2019}, where a cellular-connected UAV has to navigate from a start to an end position maintaining connectivity with a ground network exploiting a radio map. The approach includes global radio map compression to reduce computational complexity but is not based on RL and includes no higher precision local map or hard navigation constraints. To the best of our knowledge, no dual global-local map processing method applicable to multiple mission types for autonomous UAVs has been suggested previously.

The main contributions of this paper are the following:
\begin{itemize}
    \item Establishing the presented DRL approach as a general method for UAV path planning by demonstrating its applicability to two distinctly different mission scenarios: coverage path planning and path planning for wireless data harvesting;
    \item Introducing a novel approach\footnote{https://www.github.com/theilem/uavSim.git} to exploit global-local map information that allows DRL for path planning to scale to large, realistic scenario environments efficiently, with an order of magnitude more grid cells compared to earlier works \cite{Theile2020,Bayerlein2020};
    \item Overcoming the limitation of fixed target zones in previous DRL CPP approaches \cite{Theile2020} by extending control policy generalization over scenario parameters to randomly generated target zones;
    \item Analyzing and discussing the effects of key map processing parameters on the path learning performance.
\end{itemize}

\section{Problem Formulation}
\label{sec:problem}
In the following, we show that a universal problem description of coverage path planning and path planning for data harvesting can be established through separation into two parts: the environment and the target.

\subsection{Environment and UAV Model}
We consider a square grid world of size $M \times M \in \mathbb{N}^2$ with cell size $c$, where $\mathbb{N}$ is the set of natural numbers. The environment contains designated start/landing positions, regulatory no-fly zones (NFZs), and obstacles. The map can be described through a tensor $\mathbf{M}\in\mathbb{B}^{M\times M\times 3}$, where $\mathbb{B} = \{0, 1\}$ and with the start/landing zones in map-layer 1, the union of NFZs and obstacles in map-layer 2, and the obstacles alone in map-layer 3.

The UAV moves through this environment at a constant altitude $h$ occupying one cell of the environment. Its position can thus be defined through  $\mathbf{p}(t) \in \mathbb{N}^2$. The movement of the UAV is constrained through collision avoidance with obstacles and not entering NFZs. Additionally, the UAV must start and end its mission in any cell belonging to the start and landing zones while staying within its maximum flying time determined by its initial battery level. The battery level of the UAV $b(t)$ is set to $b_0 \in \mathbb{N}$ at time $t=0$ and is decremented by 1 per action step.

\subsection{Target and Mission Definitions}
\subsubsection{Coverage Path Planning}
In coverage path planning, the mission is to cover a designated target area by flying above or near it, such that it is in the field of view of a camera-like sensor mounted underneath the UAV. The target area can be described through $\mathbf{T}(t) \in \mathbb{B}^{M\times M}$, in which each element describes whether a cell has to be covered or not. The current field of view of the camera can be described with $\mathbf{V}(t) \in \mathbb{B}^{M\times M}$ indicating for each cell whether it is in the current field of view or not. In this work, the field of view is a square of $5\times 5$ surrounding the current UAV position. Additionally, buildings can block line-of-sight, which is also incorporated in calculating $\mathbf{V}(t)$. This prohibits the UAV from \textit{seeing} around the corner. 

Consequently, the target area evolves according to 
\begin{equation}
    \mathbf{T}(t+1) = \mathbf{T}(t) \land \lnot \mathbf{V}(t),
    \label{eq:coverage_target}
\end{equation}
in which $\land$ and $\lnot$ are the cell-wise logical \textit{and} and \textit{negation} operators, respectively. In our mission definition, obstacle cells in the environment cannot be a coverage target, while start and landing zones and no-fly-zones can be. The goal is to cover as much of the target area as possible within the maximum flying time constraint.

\subsubsection{Data Harvesting}
Conversely, the mission in path planning for wireless data harvesting is to collect data from $K\in\mathbb{N}$ stationary IoT devices spread throughout the environment at ground-level, with the position of device $k \in [1,K]$ given through $\mathbf{u}_k \in \mathbb{N}^2$. Each device has an amount of data $D_k(t)\in\mathbb{R}$ to be collected by the UAV. The data throughput $C_k(t)$ between the selected device $k$ and the UAV is based on the standard log-distance path loss model with Gaussian shadow fading and whether they can establish a line-of-sight connection or are obstructed by obstacles. The UAV is communicating with one device at a time and selects the device with remaining data and the highest possible data rate. A detailed description of the link performance and multiple access protocol can be found in \cite{Bayerlein2020}. The data at each device evolves according to
\begin{equation}
    D_k(t + 1) = D_k(t) - C_k(t)
    \label{eq:dh_target}
\end{equation}
Devices can be located in every cell except for the starting and landing zones or inside obstacles. The goal of the data harvesting problem is to collect as much of the devices' data as possible within the maximum flying time.

\subsubsection{Unifying Map-Layer Description}
Both problems can be described through a single target map-layer $\mathbf{D}(t) \in \mathbb{R}^{M\times M}$. In CPP, the target map-layer is given through $\mathbf{T}(t)$ evolving according to \eqref{eq:coverage_target}. In DH, the target map-layer shows the amount of available data in each cell that one of the devices is occupying, i.e. the cell at position $\mathbf{u}_k$ has value $D_k(t)$ and is evolving according to \eqref{eq:dh_target}. If a cell does not contain a device or the device data has been collected fully, the cell's value is 0. Since the two problems can be described with similar state representations, both can be solved through deep reinforcement learning with a neural network having the same structure.

\section{Methodology}
\label{sec:methodology}

While a variety of methods exist to solve the CPP and DH problems separately, the approach presented in the following can be directly applied to both distinct path planning problems. In most classical CPP approaches, individual target zones are extracted through segmentation and then connected with distance costs into a graph, while each segment is covered with a boustrophedon path. This reduces the CPP problem to an instance of the travelling salesman problem (TSP), which is NP-hard and can be solved by classical methods, e.g. as demonstrated in \cite{Xie2019} at the price of an exponential increase in time complexity with the number of target zones. 

In principle, the DH problem can also be converted into a TSP with the IoT devices as nodes in the graph and the distances between the devices as edge costs. However, the conversion neglects that communication with the device happens while traveling to and from it. In general, the optimal behavior in DH problems is not a sequential visit of all devices, as data can already be efficiently collected by establishing a LoS link from farther away, or a large amount of data waiting to be collected might require the drone to hover for an extended period of time near the device. These constraints in conjunction with stochastic communication channel models and the various possibilities for the choice of multiple access protocol are non-trivial to model and solve with classical approaches. For both problems, the UAV battery constraint adds another complication for classical approaches, as full coverage or full collection are not always feasible. The following DRL methodology allows us to combine all goals and constraints of the respective path planning problems directly without the need for additional approximations.

\subsection{Partially Observable Markov Decision Process}
To address the aforementioned problems we formulate them as a partially observable Markov decision process (POMDP)\cite{Kaelbling1998} which is defined through the tuple $(\mathcal{S}, \mathcal{A}, P, R, \Omega, \mathcal{O}, \gamma)$. In the POMDP, $\mathcal{S}$ describes the state space, $\mathcal{A}$ the action space, and $P : \mathcal{S}\times\mathcal{A}\times\mathcal{S}\mapsto\mathbb{R}$ the transition probability function. $R : \mathcal{S}\times\mathcal{A}\times\mathcal{S}\mapsto\mathbb{R}$ is the reward function mapping state, action, and next state to a real valued reward. The observation space is defined through $\Omega$ and $\mathcal{O}:\mathcal{S}\mapsto \Omega$ is the observation function. The discount factor $\gamma \in [0,1]$ varies the importance of long and short term rewards.

We unify the UAV path planning problems by describing their state space with 
\begin{equation}\label{eq:state_space}
    \mathcal{S} = 
    \underbrace{\mathbb{B}^{M\times M \times 3}}_{\substack{\text{Environment}\\ \text{Map}}}\times 
    \underbrace{\mathbb{R}^{M\times M}}_{\substack{\text{Target}\\ \text{Map}}}\times 
    \underbrace{\mathbb{N}^2}_{\text{Position}}\times
    \underbrace{ \mathbb{N}}_{\substack{\text{Flying}\\ \text{Time}}},
\end{equation}
in which the elements $s(t) \in \mathcal{S}$ are
\begin{equation}
    s(t) = (\mathbf{M},\mathbf{D}(t),\mathbf{p}(t), b(t)).
\end{equation}
The four components of the tuple are
\begin{itemize}
    \item $\mathbf{M}$ the environment map containing start and landing zones, no-fly zones, and obstacles;
    \item $\mathbf{D}(t)$ the target map indicating remaining data at device locations or remaining cells to be uncovered at time $t$;
    \item $\mathbf{p}(t)$ the UAV's position at time $t$;
    \item $b(t)$ the UAV's remaining movement budget at time $t$;
\end{itemize}
Action $a(t)\in\mathcal{A}$ of the UAV at time $t$ is given as one of the possible actions
\begin{equation*}
    \mathcal{A} = \{\text{north}, \text{east}, \text{south}, \text{west}, \text{hover}, \text{land}\}.
\end{equation*}
The generalized reward function $R(s(t),a(t),s(t+1))$ consists of the following elements:
\begin{itemize}
    \item $r_{c}$ \textit{(positive)} the data \underline{c}ollection or cell \underline{c}overing reward given by the collected data or the amount of newly covered target cells, comparing $s(t+1)$ and $s(t)$;
    \item $r_{sc}$ \textit{(negative)} safety controller (SC) penalty in case the drone has to be prevented from colliding with a building or entering an NFZ;
    \item $r_{mov}$ \textit{(negative)} constant movement penalty that is applied for every action the UAV takes without completing the mission;
    \item $r_{crash}$ \textit{(negative)} penalty in case the drone's remaining flying time reaches zero without having landed safely in a landing zone.
\end{itemize}
\begin{figure*}
    \centering
    \vspace{5pt}
    \includegraphics[width=0.95\textwidth]{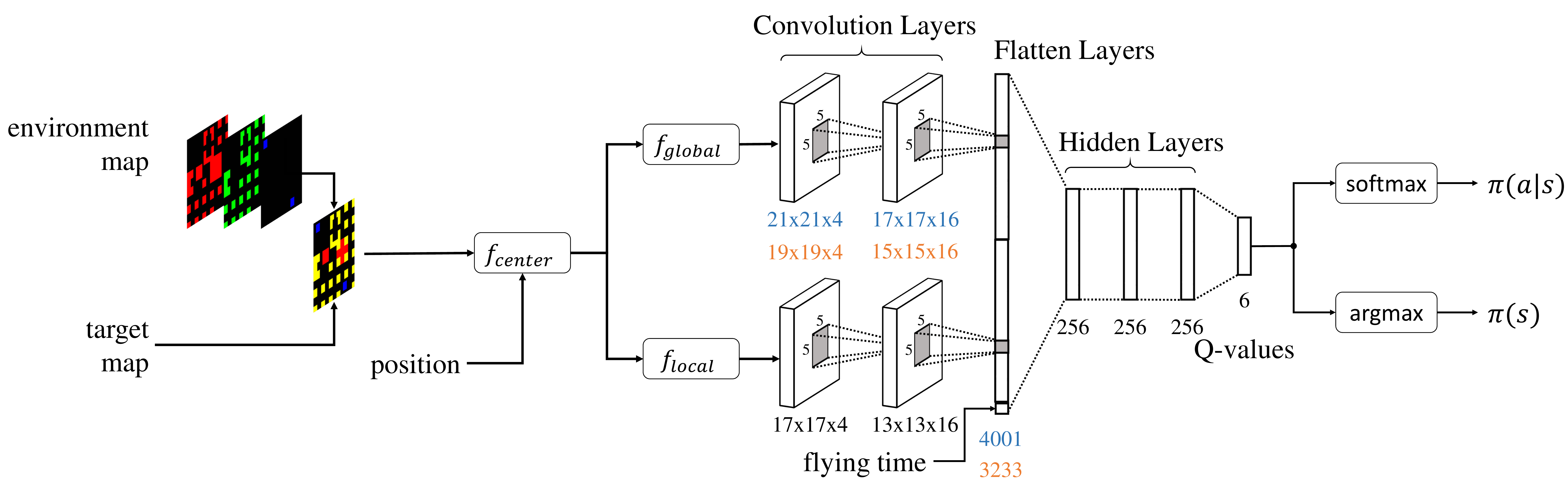}
    \caption{DQN architecture with map centering and global and local mapping, showing differences of layer size in blue for the 'Manhattan32' and orange for the 'Urban50' scenario.}
    \label{fig:figure2}
\end{figure*}
\subsection{Map Processing}
To aid an agent in interpreting the large state space given in \eqref{eq:state_space}, two map processing steps are used. The first is centering the map around the agent's position, shown in \cite{Bayerlein2020} to improve the agent's performance significantly. The downside of this approach is that it increases the representation size of the state space even further. Thus, the second map processing step, the main contribution of this work, is to present the centered map as two inputs: a full-detail local map showing the agent's immediate surroundings and a compressed global map showing the entire environment with less detail. The mathematical description of the three functions is presented in the following. Fig. \ref{fig:figure2} indicates where the functions are used within the data pipeline.
\subsubsection{Map Centering}
Given a tensor $\mathbf{A} \in \mathbb{R}^{M\times M\times n}$ describing the map layers of the environment, a centered tensor $\mathbf{B} \in \mathbb{R}^{M_c\times M_c\times n}$ with $M_c = 2M-1$ is defined through 
\begin{equation}
    \mathbf{B} = f_\text{center}(\mathbf{A}, \mathbf{p}, \mathbf{x}_{\text{pad}}), 
\end{equation}
with the centering function defined as
\begin{equation}
    f_\text{center}: \mathbb{R}^{M\times M\times n} \times \mathbb{N}^2\times \mathbb{R}^n \mapsto \mathbb{R}^{M_c\times M_c\times n}.
\end{equation}
The elements of $\mathbf{B}$ with respect to the elements of $\mathbf{A}$ are defined as
\begin{equation}\small
\mathbf{b}_{i,j} = 
    \begin{cases}
    \mathbf{a}_{i+p_0-M+1, j+p_1-M+1}, &\phantom{\land~}M \leq i+p_0+1 < 2M\\
    & \land~ M \leq j+p_1+1 < 2M\\
    \mathbf{x}_{\text{pad}}, & \text{otherwise},
    \end{cases}
\end{equation}
effectively padding the map layers of $\mathbf{A}$ with the padding value $\mathbf{x}_\text{pad}$. Note that $\mathbf{a}_{i,j}$, $\mathbf{b}_{i,j}$, and $\mathbf{x}_\text{pad}$ are vector valued of dimension $\mathbb{R}^n$. For both problems, the map layers are padded with $[0,1,1,0]^{\operatorname{T}}$, i.e. NFZs and obstacles. A qualitative description of centering with an example can be found in \cite{Bayerlein2020}.

\subsubsection{Global-Local Mapping}
The tensor $\mathbf{B} \in \mathbb{R}^{M_c\times M_c\times n}$ resulting from the map centering function is processed in two ways. The first is creating a local map according to
\begin{equation}
    \mathbf{X} = f_\text{local}(\mathbf{B},l)
\end{equation}
with the local map function defined by
\begin{equation}
    f_\text{local}: \mathbb{R}^{M_c\times M_c\times n} \times \mathbb{N} \mapsto \mathbb{R}^{l\times l\times n}.
\end{equation}
The elements of $\mathbf{X}$ with respect to the elements of $\mathbf{B}$ are defined as
\begin{equation}
\mathbf{x}_{i,j} = \mathbf{b}_{i+M-\lceil\frac{l}{2} \rceil, j+M-\lceil\frac{l}{2} \rceil}.
\end{equation}
This operation is effectively a central crop of size $l\times l$.

The global map is created according to
\begin{equation}
    \mathbf{Y} = f_\text{global}(\mathbf{B},g)
\end{equation}
with the global map function defined by
\begin{equation}
    f_\text{global}: \mathbb{R}^{M_c\times M_c\times n} \times \mathbb{N} \mapsto \mathbb{R}^{\lfloor\frac{M_c}{g} \rfloor\times \lfloor\frac{M_c}{g} \rfloor\times n}.
\end{equation}
The elements of $\mathbf{Y}$ with respect to the elements of $\mathbf{B}$ are defined as
\begin{equation}
\mathbf{y}_{i,j} = \frac{1}{g^2} \sum_{u=0}^{g-1}\sum_{v=0}^{g-1} \mathbf{b}_{gi+u, gj+v},
\end{equation}
which is an operation equal to average pooling. 

The functions $f_\text{local}$ and $f_\text{global}$ are parameterized through $l$ and $g$, respectively. Increasing $l$ increases the size of the local map, whereas increasing $g$ increases the size of the average pooling cells, decreasing the size of the global map.

\subsection{Observation Space}
The observation space $\Omega$, which is the input to the agent, is given as
\begin{equation*}
    \Omega = \Omega_l\times\Omega_g \times \mathbb{N}
\end{equation*}
containing the local map
$
    \Omega_l = \mathbb{B}^{l\times l\times 3}\times \mathbb{R}^{l\times l}
$
and the global map
$
    \Omega_g = \mathbb{R}^{\lfloor\frac{M_c}{g}\rfloor\times \lfloor\frac{M_c}{g}\rfloor\times 3}\times \mathbb{R}^{\lfloor\frac{M_c}{g}\rfloor\times \lfloor\frac{M_c}{g}\rfloor}.
$
Note that the compression of the map layers through average pooling transforms the environment layers from boolean to real. The observations $o(t) \in \Omega$ are defined through the tuple
\begin{equation}
    o(t) = (\mathbf{M}_{l}(t), \mathbf{D}_{l}(t), \mathbf{M}_{g}(t), \mathbf{D}_{g}(t), b(t)).
\end{equation}
In the observation, $\mathbf{M}_{l}(t)$ and $\mathbf{M}_{g}(t)$ are the local and global observations of the environment, and $\mathbf{D}_{l}(t)$ and $\mathbf{D}_{g}(t)$ are the local and global observations of the target, respectively. $b(t)$ is the remaining flying time of the UAV and is equal to the one in the state space. Note that the local and global observations of the environment are time-dependent, as they are centered around the time-dependent position of the UAV.

The mapping from state to observation space is given by
$
    \mathcal{O} : \mathcal{S}\mapsto \Omega
$,
with the elements $o(t)\in\mathcal{O}$ defined as:
\begin{subequations}\label{eq:observation}
\begin{align}
    \mathbf{M}_{l}(t) =& f_\text{local}(f_\text{center}(\mathbf{M},\mathbf{p}(t),[0,1,1]^\text{T}), l)\\
    \mathbf{D}_{l}(t) =& f_\text{local}(f_\text{center}(\mathbf{D}(t),\mathbf{p}(t),0), l)\\
    \mathbf{M}_{g}(t) =& f_\text{global}(f_\text{center}(\mathbf{M},\mathbf{p}(t),[0,1,1]^\text{T}), g)\\
    \mathbf{D}_{g}(t) =& f_\text{global}(f_\text{center}(\mathbf{D}(t),\mathbf{p}(t),0), g)
\end{align}
\end{subequations}

By feeding the observation space $\Omega$ into the agent instead of the state space $\mathcal{S}$, the problem is artificially converted into a partially observable MDP. The partial observability results from the restricted size of the local map and the averaging in the global map. With the following results, we show that partial observability does not make the problem infeasible for memory-less agents and that the compression greatly reduces the size of the neural network, yielding significantly less training time.

\subsection{Double Deep Reinforcement Learning - Neural Network}

To solve the aforementioned POMDP, we use reinforcement learning, specifically double deep Q-networks (DDQNs) proposed by Van Hasselt \textit{et al.} \cite{VanHasselt2016}. DDQNs approximate the Q-value of each state-action pair given as
\begin{equation}\small
    Q^\pi(s(t),a(t)) = \mathbb{E}_{\pi} \left[\sum_{k=t}^T \gamma^{k-t} R(s(k),a(k), s(k+1))\right],
    \label{eq:q}
\end{equation}
describing the discounted cumulative reward of an agent following policy $\pi$. To converge to the optimal Q-value the agent explores the environment, collecting experiences $(s(t), a(t), r(t), s(t+1))$ and storing them as $(s, a, r, s^\prime)$ in a replay memory $\mathcal{D}$, omitting temporal information. Two Q-networks parameterized through $\theta$ and $\bar{\theta}$ are used, in which the first Q-network is updated by minimizing the loss
\begin{equation}
    L(\theta) = \mathbb{E}_{s,a,s^\prime \sim \mathcal{D}}[(Q_\theta(s,a) - Y(s,a,s^\prime))^2]
    \label{eq:loss}
\end{equation}
given by experiences in the replay memory. The target value is given by 
\begin{equation}
    Y(s,a,s^\prime) = r(s,a) + \gamma Q_{\bar{\theta}}(s^\prime, \argmax_{a^\prime}Q_{\theta}(s^\prime, a^\prime)).
    \label{eq:target}
\end{equation}
The parameters of the second Q-network are updated as $\bar{\theta} \leftarrow (1-\tau)\bar{\theta} + \tau\theta$ with the soft update parameter $\tau \in (0,1]$. To address training sensitivity to the size of the replay memory we make use of combined experience replay proposed by Zhang and Sutton \cite{Zhang2017}. 


The neural network architecture used for both Q-networks is shown in Fig. \ref{fig:figure2}. The environment map and target map are stacked and centered around the UAV position and then converted into global and local observation components. After being fed through two convolutional layers each, the resulting tensors are flattened and concatenated with the remaining flying time input and passed through three hidden layers with ReLU activation functions. The output layer with no activation function represents the Q-values directly, passed through a softmax function to create an action distribution for exploration or an argmax function for exploitation.

The relevant parameter for scalability is the size of the flatten layer. It can be calculated through
\begin{equation}\label{eq:flatten_size}\small
    N = n_k \left(\left(l - n_c \lfloor\frac{s_k}{2}\rfloor\right)^2 + \left(\lfloor\frac{M_c}{g}\rfloor -  n_c \lfloor\frac{s_k}{2}\rfloor\right)^2\right) + 1
\end{equation}
with $n_k$ being the number of kernels, $n_c$ the number of convolutional layers, and $s_k$ being the kernel size. Setting the global map scaling parameter to $g=1$ and the local map size to $l=0$ deactivates global-local map processing, i.e., no downsampling and no extra local map. The parameters used in evaluation are listed in Table \ref{table:parameters}.

\begin{table}
\center
\small
\vspace{5pt}
\begin{tabular*}{\columnwidth}{cccl}
\toprule[1.5pt]
Parameter & $32 \times 32$ & $50 \times 50$ & Description\\
\midrule
$|\theta|$ & 1,175,302 &  978,694 & trainable parameters\\
$l$ & 17  & 17 & local map size\\
$g$ & 3  & 5 & global map scaling\\
$n_c$ & \multicolumn{2}{c}{2} & number of conv. layers\\
$n_k$ & \multicolumn{2}{c}{16} & number of kernels\\
$s_k$ & \multicolumn{2}{c}{5} & conv. kernel width\\
\bottomrule[1.5pt]
\end{tabular*}
\caption{Hyperparameters for $32 \times 32$ and $50 \times 50$ maps.}
\label{table:parameters}
\end{table}
\section{Simulations}
\label{sec:simulations}
\begin{figure*}[h]
    \centering
    \begin{subfigure}[t]{0.25\textwidth}
        \centering
        \includegraphics[width=\textwidth]{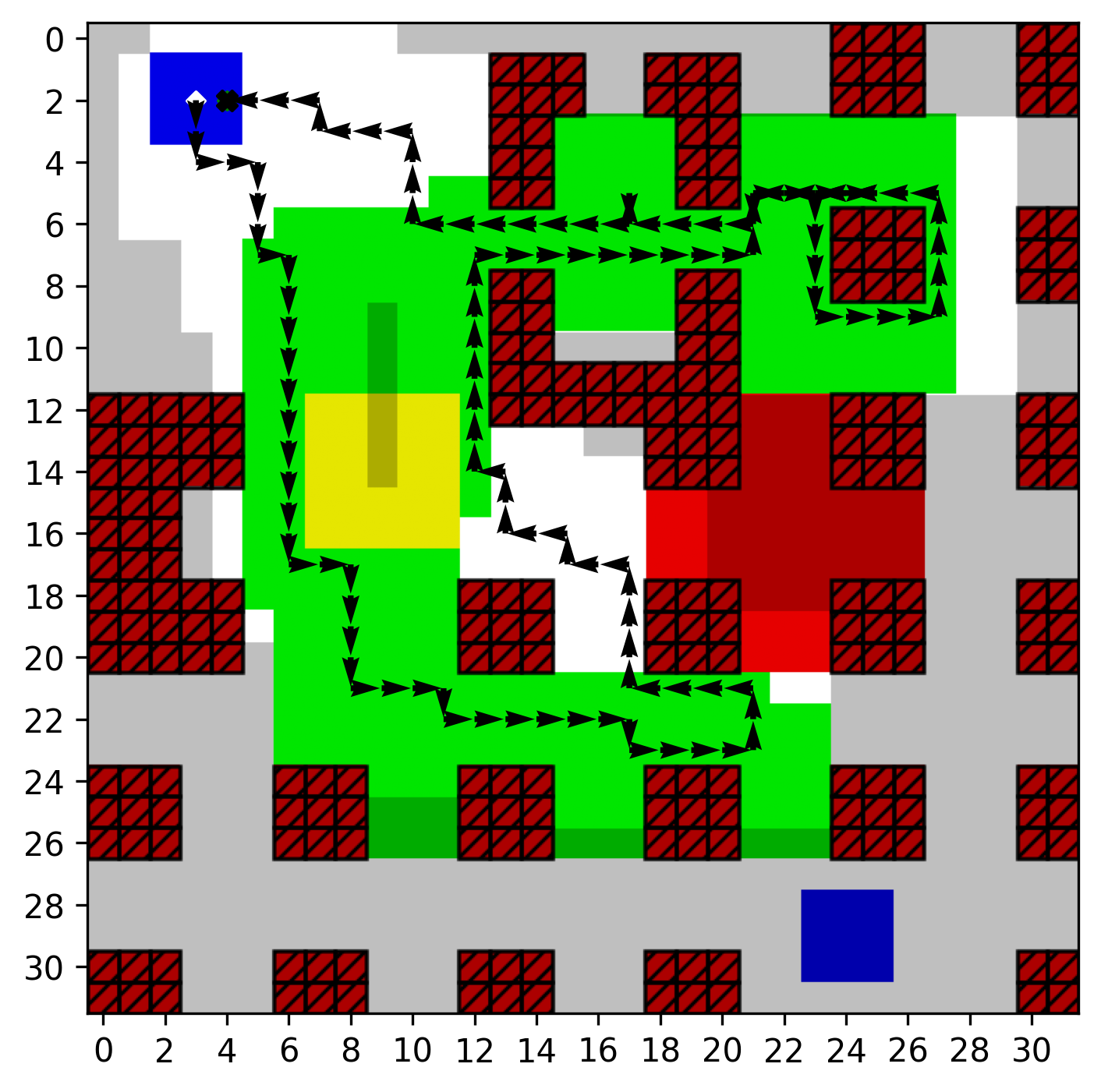}
        \caption{Movement 124/140, CR=0.94}
        \label{fig:icra_cov}
    \end{subfigure}%
    \begin{subfigure}[t]{0.25\textwidth}
        \centering
        \includegraphics[width=\textwidth]{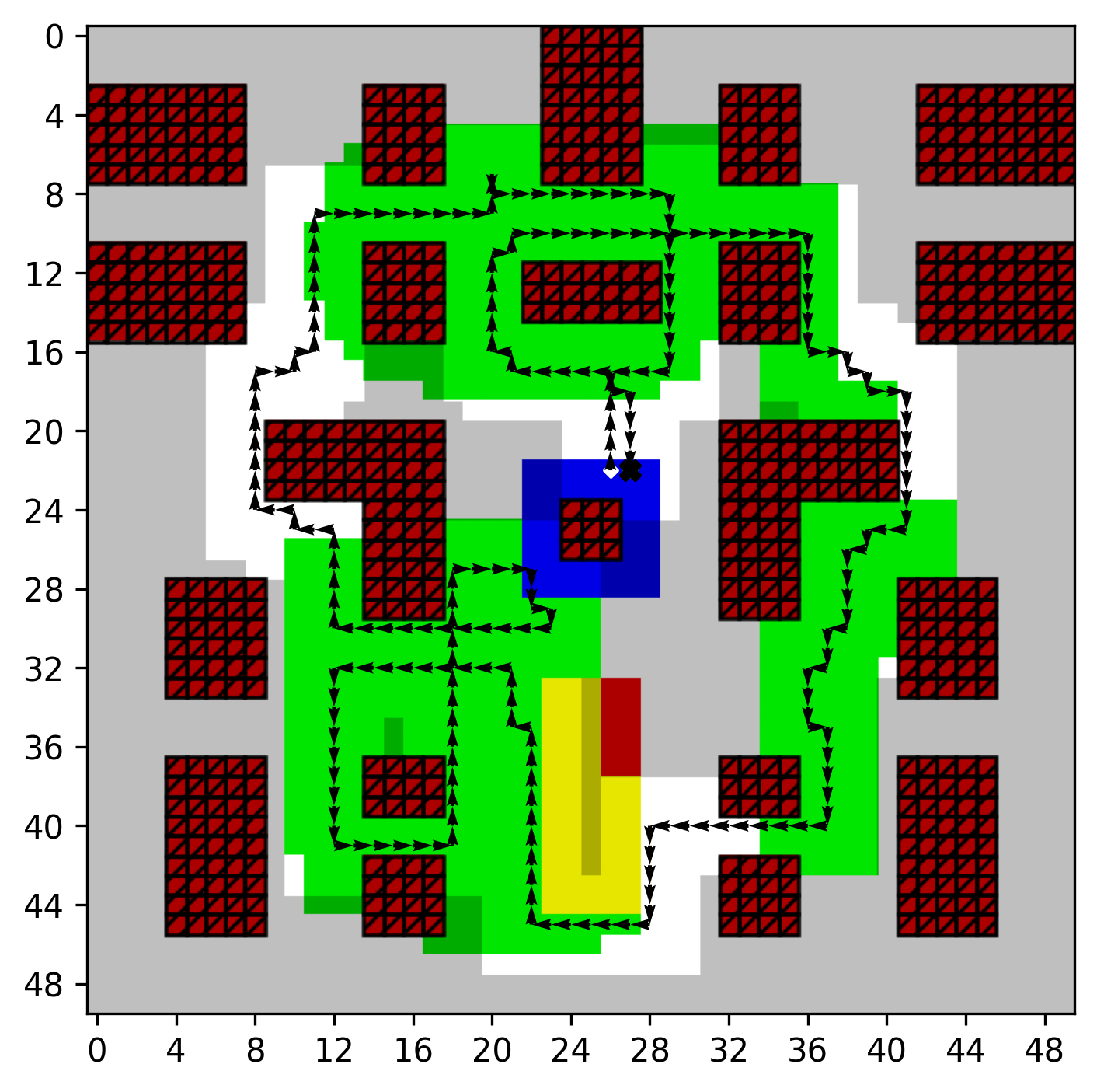}
        \caption{Movement 234/250, CR=0.94}
        \label{fig:icra_50_cov}
    \end{subfigure}%
    \begin{subfigure}[t]{0.25\textwidth}
        \centering
        \includegraphics[width=0.98\textwidth]{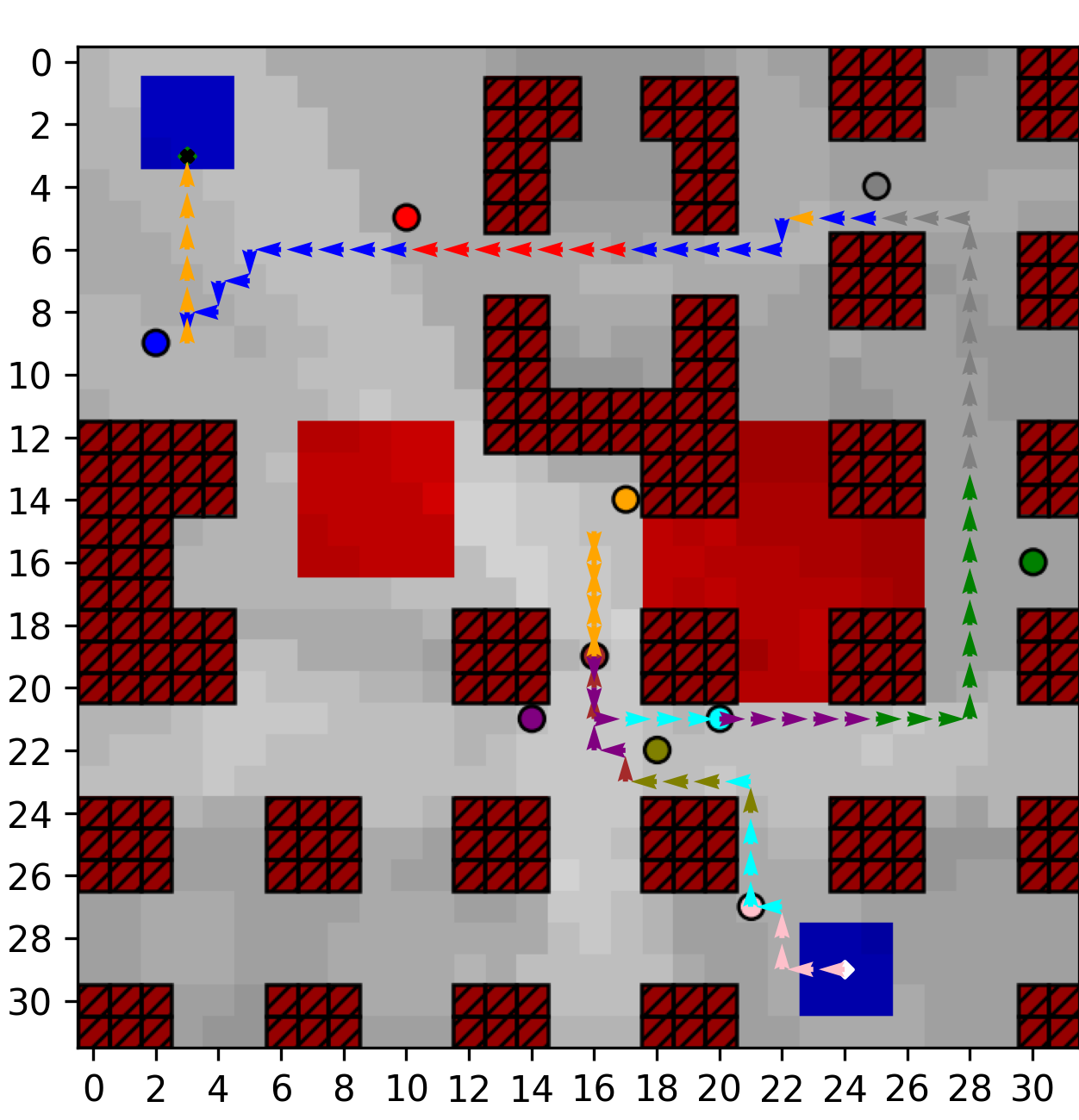}
        \caption{Movement 92/150, CR=0.99}
        \label{fig:icra_iot}
    \end{subfigure}%
    \begin{subfigure}[t]{0.25\textwidth}
        \centering
        \includegraphics[width=0.98\textwidth]{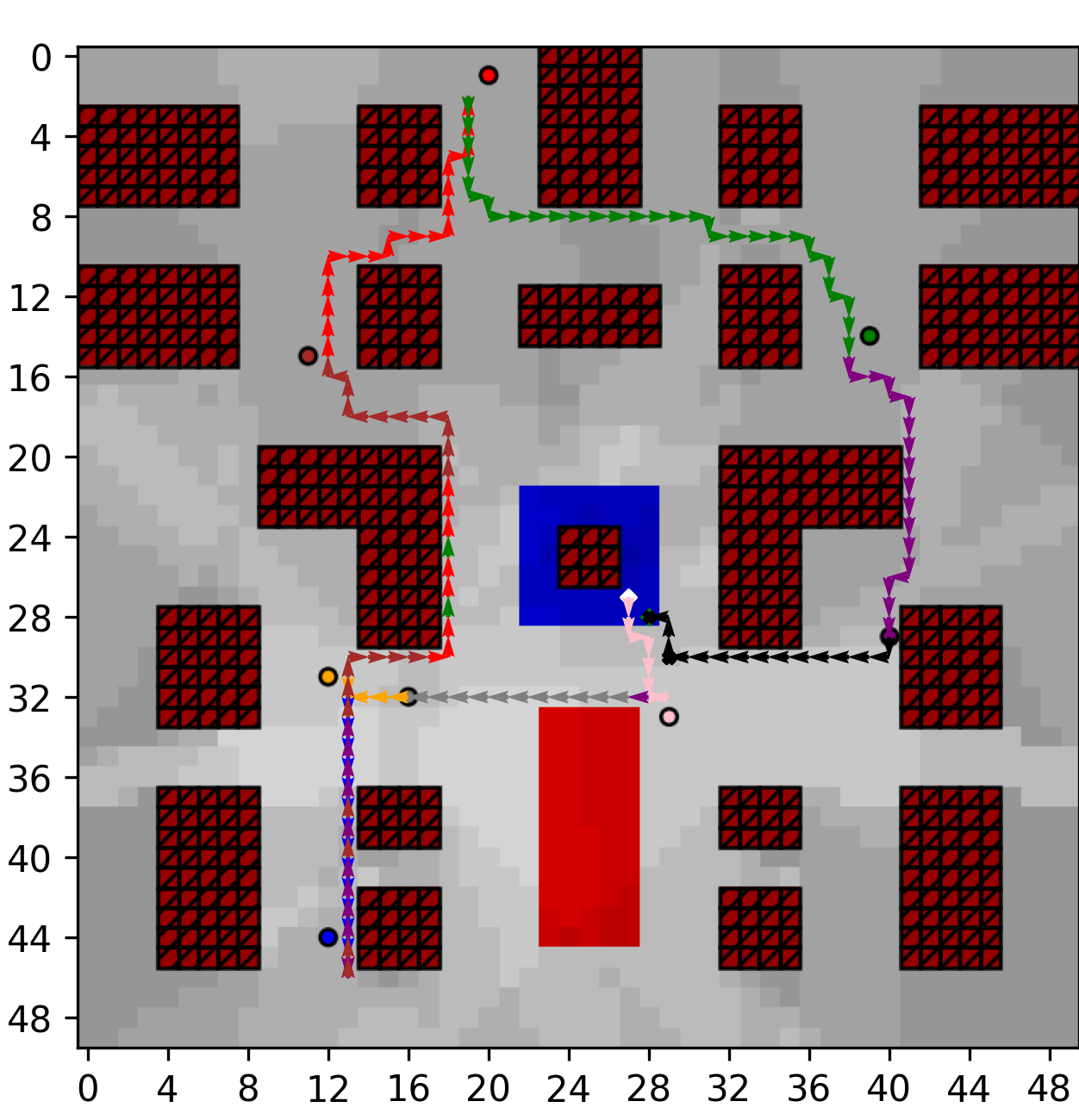}
        \caption{Movement 191/200, CR=1.0}
        \label{fig:icra_50_iot}
    \end{subfigure}
    \caption{Example trajectories from the Monte Carlo simulations for CPP (a)+(b) and DH (c)+(d) on $32\times 32$ Manhattan map (a)+(c) and $50 \times 50$ Urban map (b)+(d).}
    \label{fig:icra_cov_iot}
    \vspace{-10pt}
\end{figure*}
\subsection{Simulation Setup}
The UAV is flying in two different grid worlds. The 'Manhattan32' scenario (Fig. \ref{fig:icra_cov} and \ref{fig:icra_iot}) with $32\times 32$ cells with two starting and landing zones in the top left and bottom right corners. Besides regular building patterns, some irregularly shaped buildings and additional NFZs are present. The 'Urban50' scenario (Fig. \ref{fig:icra_50_cov} and \ref{fig:icra_50_iot}) contains $50\times 50$ cells and one starting and landing area around the center building. Buildings are generally larger and spaced out, and an additional large NFZ is present on the bottom of the map. Note that the number of cells in the 'Urban50' map is roughly one magnitude larger than in the previous works \cite{Theile2020} and \cite{Bayerlein2020}. The cell size for the scenarios is $10\si{m} \times 10\si{m}$ with Table \ref{table:legend} providing a legend for the plots. 
\subsubsection{Coverage Path Planning}
For the CPP problem, the UAV is flying at a constant altitude of $25\si{m}$ with a camera mounted underneath that has a field of view angle of $90\degree$. Consequently, the UAV can cover an area of $5\times 5$ cells simultaneously, as long as obstacles do not block line of sight. The target areas are generated by randomly sampling geometric shapes of different sizes and types and overlaying them, creating partially connected target zones. For evaluation, a traditional metric for the CPP problem is the path length. However, this metric only offers meaningful comparison when full coverage is possible. In this work, we investigate flight time constrained CPP, in which full coverage is often impossible. Therefore, the evaluation metrics used are the coverage ratio (CR), i.e. the ratio of covered target cells to the initial target cells at the end of the episode, and coverage ratio and landed (CRAL), which is zero if the UAV did not land successfully and equal to CR if it did. The benefits of the CRAL metric are that it combines the two goals, achieving high coverage and returning to the landing zone within the flight time constraint. By normalizing performance to a value in $[0,1]$, it enables performance comparisons over the changing scenarios with randomly generated target zones. 
\subsubsection{Data Harvesting}
In the DH problem, the UAV is flying at a constant altitude of $10\si{m}$ communicating with devices on ground level. The achievable data rate is calculated based on distance, random shadow fading, and line-of-sight condition with the same communication channel parameters used in \cite{Bayerlein2020}. As in CPP, the path length is not an applicable metric. It is impossible to collect all data in all scenarios depending on the randomly changing locations of IoT devices, data amount, and maximum flying time. Therefore, the evaluation metric used is the collection ratio (CR), describing the ratio of collected data from all devices to the initially available data summed over all devices. Like in CPP, we also use collection ratio and landed (CRAL) in this context, showing the full data collection and landing performance in one normalized metric.

\subsection{General Evaluation}

\begin{table}
\center
\small
\vspace{5pt}
\begin{tabular*}{\columnwidth}{lcl}
\toprule[1.5pt]
&Symbol & Description\\
\midrule
\multirow{5}{*}{\rotatebox[origin=c]{90}{\footnotesize{DQN Input}}}
&\includegraphics[align=c,height=.3cm]{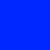} & Start and landing zone\\
&\includegraphics[align=c,height=.3cm]{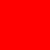} & Regulatory no-fly zone (NFZ)\\
&\includegraphics[align=c,height=.3cm]{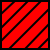} & Buildings blocking wireless links and FoV\\
&\includegraphics[align=c,height=.3cm]{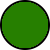} & \textbf{DH:} IoT device\\
&\includegraphics[align=c,height=.3cm]{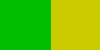} & \textbf{CPP:} Remaining target zone (yellow also NFZ)\\\midrule
\multirow{6}{*}{\rotatebox[origin=c]{90}{\footnotesize{Visualization}}}
&\includegraphics[align=c,height=.3cm]{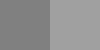} & \textbf{DH:} Summation of building shadows\\
&\includegraphics[align=c,height=.3cm]{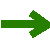} & \textbf{DH:} Movement while comm. with  green device\\
&\includegraphics[align=c,height=.3cm]{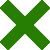} & \textbf{DH:} Hovering while comm. with green device\\
&\includegraphics[align=c,height=.3cm]{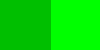} & \textbf{CPP:} Not covered and covered\\
&\includegraphics[align=c,height=.3cm]{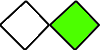} & Starting and landing positions during an episode\\
&\includegraphics[align=c,height=.3cm]{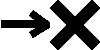} & Actions without comm.\\
\bottomrule[1.5pt]
\end{tabular*}
\caption{Legend for scenario plots, \textbf{DH} and \textbf{CPP} are only applicable in data harvesting and coverage path planning scenarios, respectively.}
\label{table:legend}
\vspace{-10pt}
\end{table}

The CPP agents were trained on target zones containing 3-8 shapes covering 20-50\% of the available area. The movement range was set to 50-150 steps for the 'Manhattan32' scenario and 150-250 for the 'Urban50' scenario. For the DH scenarios, 3-10 devices are placed randomly in free cells and contain 5.0-20.0 data units. The movement range was set to 50-150 steps for the 'Manhattan32' scenario and 100-200 for the 'Urban50' scenario. Four scenarios are evaluated in detail.

In the CPP scenarios, the agents in Fig. \ref{fig:icra_cov} and \ref{fig:icra_50_cov} show that they can find trajectories to cover most of the target area. Even the area in the NFZs is mostly covered. It can be seen that small areas that would require a detour are ignored, leading to incomplete coverage. However, most of the target area is covered efficiently.

The agents in the DH scenarios in Fig. \ref{fig:icra_iot} and \ref{fig:icra_50_iot} perform very well. In the 'Manhattan32' scenario, the agent leaves small amounts of data at the orange and purple devices totaling a collection ratio of 99.1\%. However, the agent finds a concise path, using only 92 of the allowed 150 movement steps. In the 'Urban50' scenario, the agent manages to collect all the data and return with some movement steps in spare.

All four agents were trained for 2 million steps. When analyzing their performance in all four missions using 1000 Monte Carlo generated scenarios (see Table \ref{table:metrics}), it can be seen that all agents' landing performances are good, with the 'Urban50' DH agent being slightly better.

\begin{table}
\center
\small
\vspace{5pt}
\begin{tabular*}{\columnwidth}{ccccc}
\toprule[1.5pt]
Metric & Manhattan32 & Manhattan32 & Urban50 & Urban50\\
       & CPP & DH & CPP & DH\\
\midrule
Landed & 98.5\% & 98.2\% & 98.1\% & 99.5\%\\
CR     & 71.0\% & 83.6\% & 81.5\% & 74.5\%\\
CRAL   & 70.3\% & 82.5\% & 80.1\% & 74.2\%\\
\bottomrule[1.5pt]
\end{tabular*}
\caption{Performance metrics averaged over 1000 random scenario Monte Carlo iterations.}
\label{table:metrics}
\end{table}

\subsection{Global-Local Parameter Evaluation}
To establish the performance sensitivity to the new hyper-parameters, global map scaling $g$, and local map size $l$, we trained multiple agents with different parameters on the CPP and DH problems. We chose four values for $l$ and four for $g$ and trained three agents for each possible combination. Additionally, we trained three agents without the usage of global and local map processing, which is equivalent to setting $g=1$ and $l=0$. The resulting 51 agents for the CPP and DH problems were trained for 500k steps each and evaluated on 200 Monte Carlo generated scenarios. The difference to the previous evaluation is that the movement budget range was set to $150-300$.

Table \ref{table:flatten_size} shows the selected parameters and the resulting flatten layer size according to \eqref{eq:flatten_size}. A significant speedup of the training process compared to agents without global and local map processing can be observed in Table \ref{table:cpp_speedup}.

\begin{table}
\small
\begin{center}
\vspace{5pt}
\begin{tabular*}{0.85\columnwidth}{c||c|c|c|c}
\toprule[1.5pt]Global map&\multicolumn{4}{c}{Local map scaling $l$}\\
scaling $g$& 9& 17& 25& 33\\\hline
2& 8,481& 9,761& 13,089& 18,465\\\hline
3& 2,721& 4,001&  7,329& 12,705\\\hline
5&   273& 1,553&  4,881& 10,257\\\hline
7&    33& 1,313&  4,641& 10,017\\
\bottomrule[1.5pt]
\end{tabular*}
\end{center}
\caption{Flatten layer size for 'Manhattan32' with different global map scaling and local map sizes; Without global-local map processing the size is 48,401 neurons.}
\label{table:flatten_size}
\end{table}

\begin{table}
\small\addtolength{\tabcolsep}{-1pt}
\begin{center}
\begin{tabular*}{\columnwidth}{c||c|c|c|c|c|c|c|c}
\toprule[1.5pt]Global&\multicolumn{8}{c}{Local map scaling $l$}\\
map& \multicolumn{2}{c|}{9}&  \multicolumn{2}{c|}{17}&  \multicolumn{2}{c|}{25}&  \multicolumn{2}{c}{33}\\\
scaling $g$& CPP&DH&CPP&DH&CPP&DH&CPP&DH\\\hline
2& 2.7&2.2 & 2.3&2.0 & 1.8&1.6 & 1.3&1.1\\\hline
3& 3.5&3.0 & 3.0&2.5 & 2.2&1.9 & 1.6&1.4\\\hline
5& 4.2&3.6 & 3.4&3.0 & 2.5&2.2 & 1.9&1.6\\\hline
7& 4.7&3.8 & 3.6&3.0 & 2.5&2.2 & 2.5&2.1\\
\bottomrule[1.5pt]
\end{tabular*}
\end{center}
\caption{Training time speedup for the CPP and DH problem relative to without global-local map processing.}
\label{table:cpp_speedup}
\end{table}

\begin{figure}
    \centering
    \vspace{2pt}
    \begin{subfigure}{\columnwidth}
        \centering
        \includegraphics[width=\textwidth]{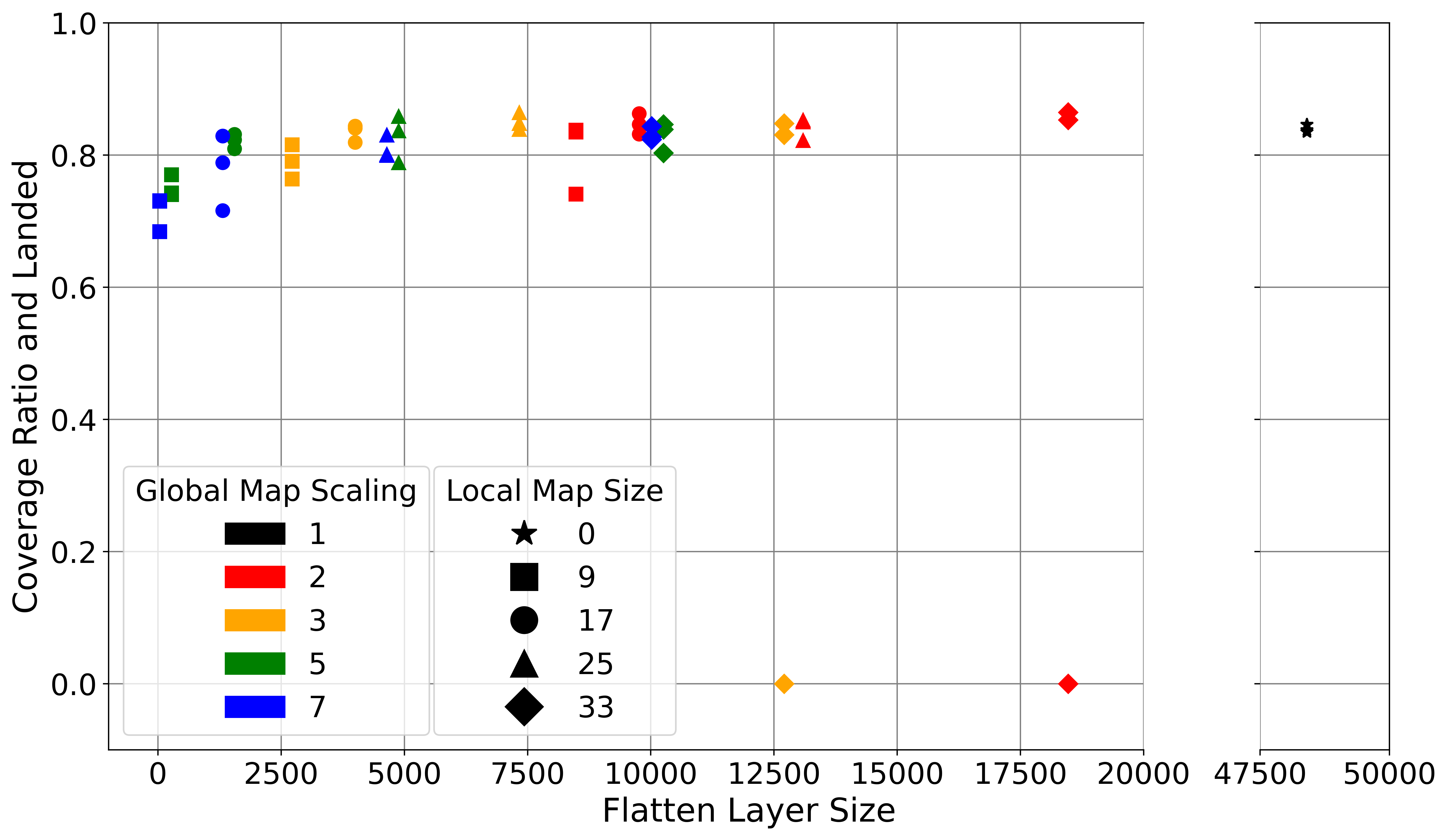}
        \caption{Grid search for CPP}
        \label{fig:cpp_grid}
    \end{subfigure}
    \begin{subfigure}{\columnwidth}
        \centering
        \includegraphics[width=\textwidth]{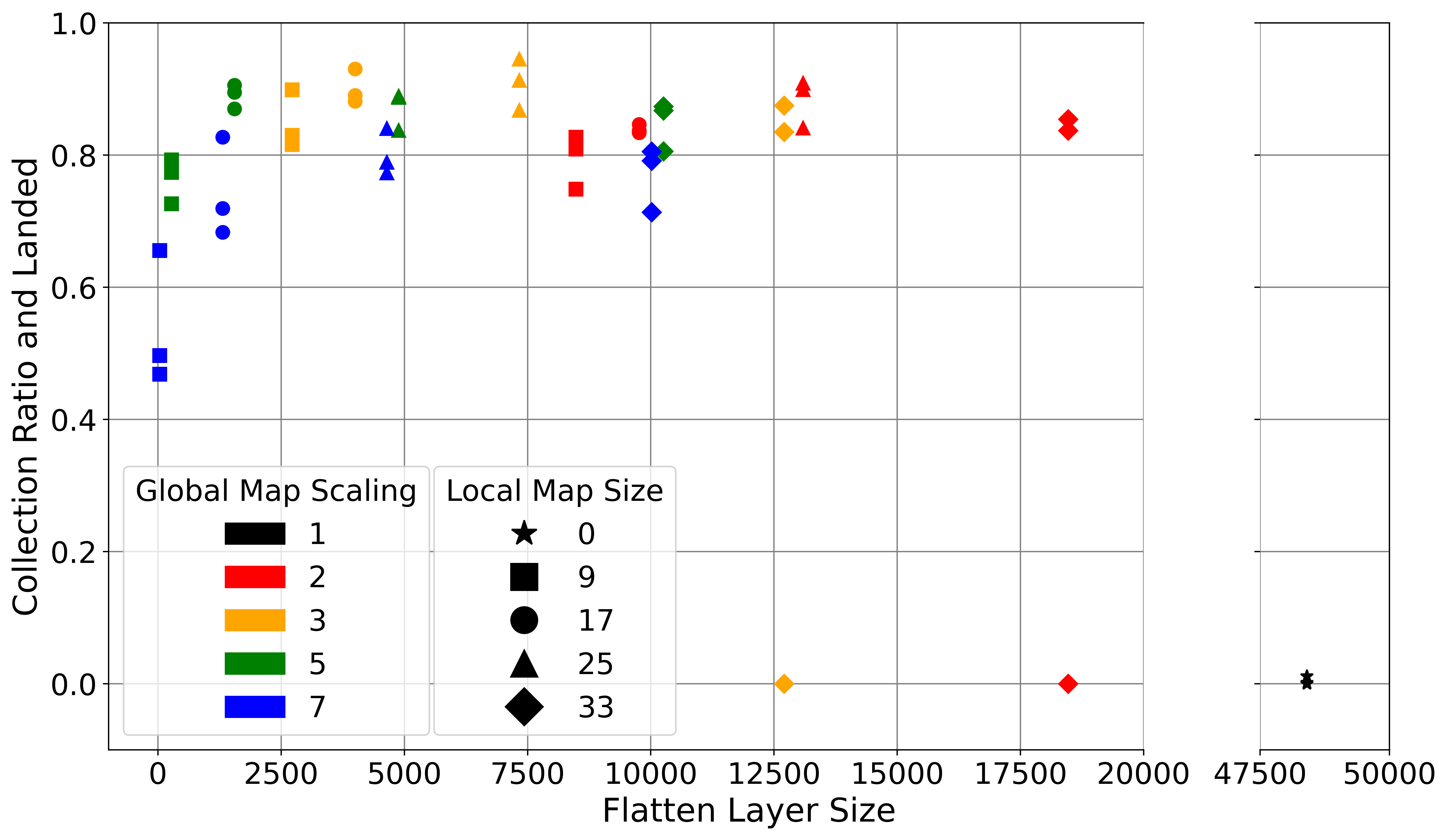}
        \caption{Grid search for DH}
        \label{fig:iot_grid}
    \end{subfigure}
    \caption{Parameter grid search for CPP and DH with parameters from Table \ref{table:flatten_size}; the black stars correspond to agents without global-local map processing.
    }
    \label{fig:grids}
\end{figure}

The resulting CRAL values from the Monte Carlo simulations for each agent with respect to the agent's flatten layer size are shown in Fig. \ref{fig:cpp_grid} and \ref{fig:iot_grid} for the CPP and DH problem, respectively. It can be seen that the DH problem is more sensitive to the parameters than the CPP problem. Generally, a larger flatten layer yields better performance up to a point. For both problems, it can be seen that a large flatten layer can cause the learning to get unstable, resulting in a CRAL of zero for some runs. This is caused by the agent's failure to learn how to land. The DH agents, which are not using the global-local map approach, never learn how to land reliably and thus have a CRAL score near zero.

In both cases, the agents with $l=17$ and $g=3$ or $g=5$ show the best performance with respect to their flatten layer size, justifying the selection in Table \ref{table:parameters}. Besides these two parameter combinations, it is noteworthy that the agents with $l=9$ and $g=7$ also perform well in both scenarios, despite their small flatten layer size of only 33 neurons.

\section{Conclusion}
\label{sec:conclusion}

We have presented a method for generalizing autonomous UAV path planning over two distinctly different mission types, coverage path planning and data harvesting. Through the flexibility afforded by combining specific mission goals and navigation constraints in the reward function, we trained DDQNs with identical structures in both scenarios to make efficient movement decisions. We have introduced a novel global-local map processing scheme that allows to feed large maps directly into convolutional layers of the DRL agent and analyzed the effects of map processing parameters on learning performance. In future work, we will investigate still existing hindrances for applying our method to even larger maps, namely avoiding small-scale decision alternation through the use of macro-actions or options \cite{Sutton1999}. Combining the presented high-level path planning approach with a low-level flight dynamics controller will also make it possible to conduct experiments with realistic open-source UAV simulators in the future. Additionally, we will investigate the effect of irregularly shaped, non-convex obstacles on the path planning performance.

\section*{Acknowledgments}
Marco Caccamo was supported by an Alexander von Humboldt Professorship endowed by the German Federal Ministry of Education and Research. Harald Bayerlein and David Gesbert were partially supported by the French government, through the 3IA Côte d’Azur project number ANR-19-P3IA-0002, as well as by the TSN CARNOT Institute under project Robots4IoT.

\balance
\bibliography{biblio}

\begin{thebibliography}{10}

\bibitem{Theile2020}
M.~Theile, H.~Bayerlein, R.~Nai, D.~Gesbert, and M.~Caccamo, ``{UAV} coverage
  path planning under varying power constraints using deep reinforcement
  learning,'' in {\em IEEE/RSJ International Conference on Intelligent Robots
  and Systems (IROS)}, 2020.

\bibitem{Bayerlein2020}
H.~Bayerlein, M.~Theile, M.~Caccamo, and D.~Gesbert, ``{UAV} path planning for
  wireless data harvesting: A deep reinforcement learning approach,'' in {\em
  IEEE Global Communications Conference (GLOBECOM)}, 2020.

\bibitem{Zeng2019}
Y.~Zeng, Q.~Wu, and R.~Zhang, ``Accessing from the sky: A tutorial on {UAV}
  communications for {5G} and beyond,'' {\em Proceedings of the IEEE},
  vol.~107, no.~12, pp.~2327--2375, 2019.

\bibitem{Shakeri2019}
R.~Shakeri, M.~A. Al-Garadi, A.~Badawy, A.~Mohamed, T.~Khattab, A.~K. Al-Ali,
  K.~A. Harras, and M.~Guizani, ``Design challenges of multi-{UAV} systems in
  cyber-physical applications: A comprehensive survey and future directions,''
  {\em IEEE Communications Surveys \& Tutorials}, vol.~21, no.~4,
  pp.~3340--3385, 2019.

\bibitem{Cabreira2019}
T.~Cabreira, L.~Brisolara, and P.~R~Ferreira, ``Survey on coverage path
  planning with unmanned aerial vehicles,'' {\em Drones}, vol.~3, no.~1, 2019.

\bibitem{Piciarelli2019}
C.~Piciarelli and G.~L. Foresti, ``Drone patrolling with reinforcement
  learning,'' in {\em Proceedings of the 13th International Conference on
  Distributed Smart Cameras}, ACM, 2019.

\bibitem{Julian2019}
K.~D. Julian and M.~J. Kochenderfer, ``Distributed wildfire surveillance with
  autonomous aircraft using deep reinforcement learning,'' {\em Journal of
  Guidance, Control, and Dynamics}, vol.~42, no.~8, pp.~1768--1778, 2019.

\bibitem{Seraj2020}
E.~Seraj and M.~Gombolay, ``Coordinated control of {UAVs} for human-centered
  active sensing of wildfires,'' in {\em American Control Conference (ACC)},
  pp.~1845--1852, IEEE, 2020.

\bibitem{Baldazo2019}
D.~Baldazo, J.~Parras, and S.~Zazo, ``Decentralized multi-agent deep
  reinforcement learning in swarms of drones for flood monitoring,'' in {\em
  27th European Signal Processing Conference (EUSIPCO)}, IEEE, 2019.

\bibitem{Esrafilian2018}
O.~Esrafilian, R.~Gangula, and D.~Gesbert, ``Learning to communicate in
  {UAV}-aided wireless networks: Map-based approaches,'' {\em IEEE Internet of
  Things Journal}, vol.~6, no.~2, pp.~1791--1802, 2018.

\bibitem{Liu2021}
C.~H. Liu, Z.~Dai, Y.~Zhao, J.~Crowcroft, D.~O. Wu, and K.~Leung, ``Distributed
  and energy-efficient mobile crowdsensing with charging stations by deep
  reinforcement learning,'' {\em IEEE Transactions on Mobile Computing},
  vol.~20, no.~1, pp.~130--146, 2021.

\bibitem{Zhang2019}
S.~Zhang and R.~Zhang, ``Radio map based path planning for cellular-connected
  {UAV},'' in {\em IEEE Global Communications Conference (GLOBECOM)}, 2019.

\bibitem{Xie2019}
J.~Xie, L.~R.~G. Carrillo, and L.~Jin, ``An integrated traveling salesman and
  coverage path planning problem for unmanned aircraft systems,'' {\em IEEE
  Control Systems Letters}, vol.~3, no.~1, pp.~67--72, 2019.

\bibitem{Kaelbling1998}
L.~P. Kaelbling, M.~L. Littman, and A.~R. Cassandra, ``Planning and acting in
  partially observable stochastic domains,'' {\em Artificial intelligence},
  vol.~101, no.~1-2, pp.~99--134, 1998.

\bibitem{VanHasselt2016}
H.~Van~Hasselt, A.~Guez, and D.~Silver, ``Deep reinforcement learning with
  double {Q}-learning,'' in {\em Thirtieth AAAI conference on artificial
  intelligence}, pp.~2094--2100, 2016.

\bibitem{Zhang2017}
S.~Zhang and R.~S. Sutton, ``A deeper look at experience replay,'' {\em
  arXiv:1712.01275 [cs.LG]}, 2017.

\bibitem{Sutton1999}
R.~S. Sutton, D.~Precup, and S.~Singh, ``Between {MDPs} and semi-{MDPs}: A
  framework for temporal abstraction in reinforcement learning,'' {\em
  Artificial intelligence}, vol.~112, no.~1-2, pp.~181--211, 1999.

\end{thebibliography}
\bibliographystyle{ieeetr}

\end{document}